\documentclass[letterpaper,10pt,conference]{ieeeconf}

\IEEEoverridecommandlockouts
\usepackage{cite}
\usepackage{amsmath,amssymb}
\usepackage{graphicx}
\usepackage{booktabs}
\usepackage{array}
\usepackage{multirow}
\usepackage{url}
\usepackage{needspace}
\usepackage[hidelinks]{hyperref}

\graphicspath{{article/asset/}{figures/active/}{figures/}}

\title{\LARGE \bf ReF-HIL: Shaping the Critic around Human Action Neighborhoods for Efficient Human-in-the-Loop Reinforcement Learning}
\author{Shaoyin Luo$^{*}$, Song Wang$^{*}$, Shibo Xia, Tianle Zhang,\\
Zhaowei Liang, Guanghui Shen, Bin Wang, and Dan Wu$^{\dagger}$%
\thanks{This work was supported by the National Natural Science Foundation
of China under Grant 52375019.}%
\thanks{$^{*}$Shaoyin Luo and Song Wang contributed equally to this work.}%
\thanks{The authors are with the Department of Mechanical Engineering,
\mbox{Tsinghua University}, Beijing 100084, China.}%
\thanks{$^{\dagger}$Corresponding author: Dan Wu
(\texttt{wud@tsinghua.edu.cn}).}%
}

\hypersetup{
  pdftitle={ReF-HIL: Shaping the Critic around Human Action Neighborhoods for Efficient Human-in-the-Loop Reinforcement Learning},
  pdfauthor={Shaoyin Luo; Song Wang; Shibo Xia; Tianle Zhang; Zhaowei Liang; Guanghui Shen; Bin Wang; Dan Wu}
}

\begin{document}
\bstctlcite{refhil:reference-style}
\maketitle
\thispagestyle{empty}
\pagestyle{empty}

\begin{abstract}
Human-in-the-loop reinforcement learning (HIL-RL) offers a promising route to efficient training of robotic manipulation policies by combining autonomous learning with human demonstrations and online corrections. However, insufficient use of successful human experience in value learning prolongs costly real-world training, while persistent imitation penalties can limit value-driven policy improvement. To address these limitations, we propose ReF-HIL, an efficient HIL-RL framework that uses human guidance to accelerate the learning process. Human-Reference-Guided Value Shaping learns an independent value reference from successful human experience to guide online value learning, while incorporating local corrective feedback. A Human Action Fence defines a learned human-action neighborhood, allowing value-driven optimization for better performance without imitation penalties inside while constraining policy and value updates outside. Experiments on five diverse and challenging real-world manipulation tasks demonstrate improved overall learning efficiency and higher success rates compared with the evaluated baselines. Specifically, ReF-HIL reaches 90\% autonomous success in only 18--63 minutes of active training and achieves final success rates of 91.7--100\%. These results highlight the potential of human-guided reinforcement learning to acquire reliable manipulation skills efficiently in the real world.
Project website: \mbox{\urlstyle{same}\url{https://anonymous.4open.science/w/ReF-HIL-7762/}}.
\end{abstract}

\section{Introduction}
\label{sec:introduction}

Reliable manipulation across diverse objects and operating conditions remains a central challenge in robotics. Learning-based approaches address this challenge by using data to capture complex physical interactions and learn manipulation policies~\cite{ai_dynamics_review_2025,diffusion_policy}. Reinforcement learning (RL), in particular, enables robots to improve their behavior by interacting with the environment and maximizing task rewards~\cite{vpcl}. However, applying RL directly to physical robots involves costly trials, failure recovery, and environment resets, making efficient use of experience essential~\cite{tang_robot_rl_review_2025}. Human-in-the-loop reinforcement learning (HIL-RL) offers a practical way to improve learning efficiency: demonstrations provide initial experience, while online human corrections guide exploration toward successful behavior~\cite{hilserl}.

\begin{figure}[t]
    \centering
    \includegraphics[width=\columnwidth]{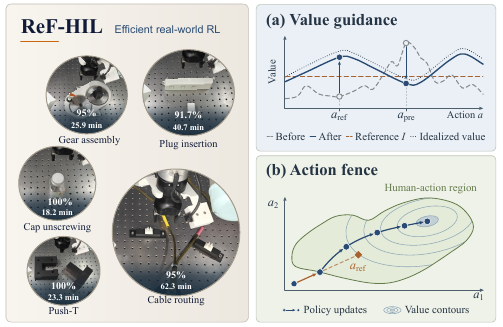}
    \caption{ReF-HIL across five diverse real-world robotic manipulation tasks. Task annotations show mean autonomous success rates from Table~\ref{tab:performance-comparison} and active training times to first reach 90\% autonomous success. (a) Successful experience supplies reference $I$; intervention feedback lowers values of preceding autonomous actions. (b) Human actions define a region for policy optimization: outside updates point toward $a_{\mathrm{ref}}$; inside updates follow value gradients without imitation penalties.}
    \label{fig:refhil-concept}
\end{figure}

However, the efficiency of HIL-RL depends on how the learner uses human interventions to improve its policy. Existing methods use interventions to shape value estimates and guide policy improvement. In value learning, intervention-based rewards and preference constraints discourage undesirable behavior and align value estimates with human judgments~\cite{haco,pvp,rlif}. Yet learning which behaviors lead to task completion requires information beyond local acceptance or rejection. If the learner does not fully exploit the value of demonstrated actions and their outcomes, it may therefore need more real-world interaction to learn effective behavior~\cite{pact,maxq}. For policy improvement, human actions serve as imitation targets, either directly or selectively, to help the policy learn feasible behavior more quickly. Since human actions may be suboptimal, persistent imitation penalties can discourage the policy from choosing better actions, potentially limiting value-driven improvement~\cite{phil,multi_hug_rl,nair_demo_rl}.

To address these limitations, we propose ReF-HIL (Fig.~\ref{fig:refhil-concept}), a framework for HIL-RL that combines Human-Reference-Guided Value Shaping with a Human Action Fence. Successful human experience provides a useful starting point, even when the human actions are suboptimal. We therefore learn an independent value reference from this experience to help prevent the online critic from undervaluing reference actions. Meanwhile, intervention feedback guides the critic to rank preceding autonomous actions below reference actions. To improve performance within limited training time, the fence focuses value-driven policy optimization on an action neighborhood learned from human demonstrations and online interventions. It removes imitation penalties inside and constrains policy and value updates outside.

The main contributions of this work are:
\begin{enumerate}
    \item We propose ReF-HIL, an efficient human-in-the-loop reinforcement learning framework that accelerates value-driven policy improvement using potentially suboptimal human guidance.
    \item We introduce Human-Reference-Guided Value Shaping to accelerate value learning with an independent reference from successful human experience, and a Human Action Fence to enable value-driven policy improvement within a learned human-action neighborhood without imitation penalties.
    \item We evaluate ReF-HIL on five diverse and challenging real-world manipulation tasks, reaching 90\% autonomous success in only 18--63 minutes of active training and achieving final success rates of 91.7--100\%. Comparisons show better overall learning efficiency and higher success rates than the evaluated baselines, while ablations and focused analyses clarify the roles of both components.
\end{enumerate}

\section{Related Work}
\label{sec:related-work}

\subsection{Human-in-the-Loop Reinforcement Learning}

Human-in-the-loop reinforcement learning uses human intervention to improve a robot's policy during online training. HIL-SERL~\cite{hilserl} combines human demonstrations and online corrections with a sample-efficient off-policy RL algorithm~\cite{rlpd}, achieving high success rates across a range of real-world manipulation tasks. However, treating these corrections only as replay data does not explicitly use the judgment behind an intervention: the human wants the robot to change its current behavior~\cite{eil}. Existing methods make further use of this feedback by penalizing behavior that leads to intervention~\cite{rlif} or assigning higher values to human actions than to the actions they replace~\cite{pvp,icopro}. The same feedback is also used to directly encourage the policy to prefer human corrections~\cite{ohp_rl} or select conservative actions to reduce intervention risk~\cite{gains}, improving learning efficiency and policy performance.

While these methods improve learning through local intervention feedback, successful human experience offers additional value guidance that can further reduce exploration. Max-Q trains its critic on recorded returns, preserving guidance from achieved outcomes. However, success achieved through later human recovery can also raise the estimated values of preceding autonomous mistakes~\cite{maxq}. PACT uses successful demonstrations to identify such suboptimal segments and lowers their value targets by comparing the online critic's estimates for human and policy actions~\cite{pact}. Yet human actions can remain undervalued even with relative ranking. ReF-HIL learns an independent reference from successful human experience to supervise online reference-action values. These values then anchor local corrections, preserving guidance from successful behavior while discouraging autonomous mistakes.

\subsection{Reinforcement Learning with Behavioral Priors}

Behavioral priors and demonstrations accelerate RL by directing exploration toward useful actions~\cite{behavior_priors,parrot} and providing supervision for learning~\cite{dqfd,dapg,phil}. However, guidance that is useful early in learning can become restrictive when further improvement requires departing from suboptimal human actions. Value filtering, advantage weighting, and uncertainty-dependent tolerances address this tension by making imitation selective or adaptive~\cite{nair_demo_rl,multi_hug_rl,awac,silri}. These mechanisms reduce reliance on human actions, but a persistent imitation penalty can still oppose beneficial deviations. This motivates using human guidance to direct learning without requiring the policy to reproduce a particular action. Corrective action sets broaden the choices admitted by human feedback~\cite{clic}, while imitation-policy candidates can guide RL execution and value backups without an imitation loss on the RL actor~\cite{ibrl}. Human behavior can thus guide learning without serving as a persistent imitation target.

Human experience can also guide where value-driven policy improvement takes place, concentrating learning near useful behavior. Such guidance can limit optimization of poorly supported actions whose values may be overestimated~\cite{cql,iql,calql}. SPOT provides this guidance through density regularization, but its preference for higher-density actions can also oppose beneficial changes within data support~\cite{spot}. EXPO applies value-driven adjustments to actions sampled from a base policy within a prescribed distance, using both the original and adjusted actions for execution and value backups~\cite{expo}. However, the distance bound can still admit actions poorly supported by human experience. Our Human Action Fence learns an action neighborhood from human demonstrations and online interventions. It constrains policy and value updates outside this neighborhood while allowing value-driven optimization without imitation penalties inside, which preserves human guidance and leaves room for further policy improvement.

\section{Problem Formulation}
\label{sec:problem-formulation}

\begin{figure*}[t]
    \centering
    \includegraphics[width=\textwidth]{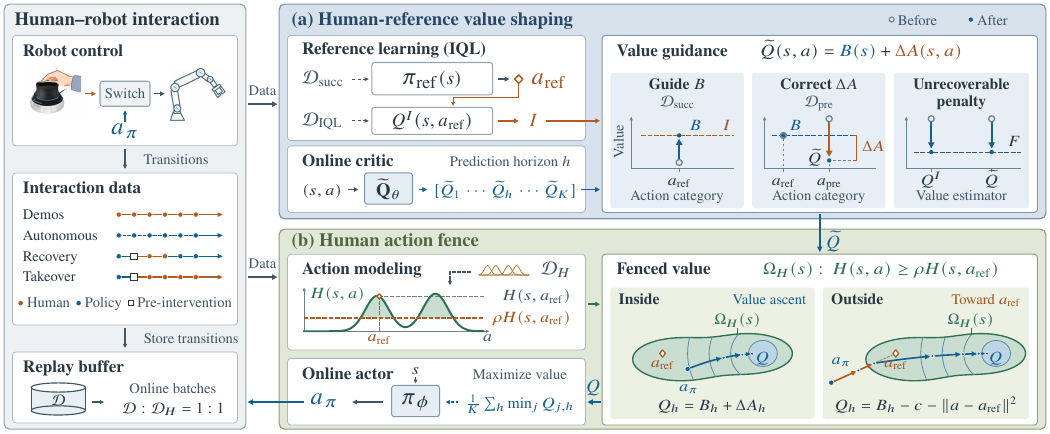}
    \caption{Overview of ReF-HIL. Human and autonomous interaction supplies the training data. (a) Evaluating the reference action $a_{\mathrm{ref}}$ with the IQL critics provides $I$, which guides the online baseline $B$. Intervention feedback corrects $\Delta A$; human-marked unrecoverable states provide target $F$ for both $Q^I$ and $\widetilde Q$. (b) Human actions train a potentially multimodal score $H$, whose relative threshold defines $\Omega_H(s)$. Updates outside this region point toward $a_{\mathrm{ref}}$, while updates inside follow value gradients. The actor maximizes the horizon average of the twin-critic minimum. Open and filled circles in (a) denote values before and after guidance.}
    \label{fig:refhil-framework}
\end{figure*}

We formulate real-world robotic manipulation as a reinforcement learning problem described by an approximate Markov decision process $\mathcal M=(\mathcal S,\mathcal A,P,r,\gamma)$. Visual and proprioceptive observations constitute the state $s\in\mathcal S$, while $\mathcal A$ and $P$ denote the robot's action space and transition dynamics. With the task reward $r$ assigning a positive value to successful completion and a step cost otherwise, the objective is to learn a deterministic policy $\pi_\phi$ that maximizes the expected discounted return under autonomous execution:
\begin{equation}
    J(\pi_\phi)=\mathbb E_{\tau\sim\pi_\phi}
    \left[\sum_{t=0}^{\infty}\gamma^t r_t\right],
    \label{eq:autonomous-objective}
\end{equation}
where $\gamma\in(0,1)$ is the discount factor and rewards are zero after successful task termination.

During training, we start with initial demonstrations and continue with autonomous interaction supported by online human interventions, without assuming that human actions are optimal. When intervening, the human may provide a brief correction before returning control to the policy or continue until task completion. The resulting experience therefore contains both human and autonomous actions; each transition $(s_t,a_t,r_t,s_{t+1})$ is recorded with an indicator $m_t\in\{0,1\}$ to distinguish human control ($m_t=1$) from autonomous control ($m_t=0$).

To improve the sample efficiency of value learning, we adopt a vector of fixed-horizon action values~\cite{fixed_horizon_td}, using the same experience to supervise return estimates over multiple prediction horizons. For a fixed deterministic policy $\pi$, this value vector is defined by
\begin{equation}
\begin{aligned}
    Q_h^\pi(s,a)
    &=\mathbb E_\pi\!\left[
        \sum_{k=0}^{h-1}\gamma^k r_{t+k}
        \,\middle|\,s_t=s,a_t=a\right],\\
    \mathbf Q^\pi(s,a)
    &=\big(Q_1^\pi(s,a),\ldots,Q_K^\pi(s,a)\big)^\top,
\end{aligned}
\label{eq:vector-value}
\end{equation}
where $h$ counts transitions from the current state-action pair up to a maximum horizon $K$, with the reward and discount factor shared across all components.

Learning targets for these horizon-specific values follow from the Bellman recursion, with $Q_0^\pi(s,a)=0$:
\begin{equation}
    Q_h^\pi(s,a)=\mathbb E\!\left[
        r+\gamma(1-d)Q_{h-1}^\pi(s',\pi(s'))
        \,\middle|\,s,a\right],
    \label{eq:horizon-recursion}
\end{equation}
where $d$ indicates successful task termination. Episode step limits are treated as truncations. The recursion builds longer-horizon targets from shorter-horizon estimates, with the one-step value grounded in the immediate reward. Both the human value reference and shaped online critics $\mathbf Q_\theta(s,a)\in\mathbb R^K$ use this representation.

\section{Method}
\label{sec:method}

\subsection{Overview}
\label{sec:method-overview}

ReF-HIL (Fig.~\ref{fig:refhil-framework}) aims to improve the efficiency of HIL-RL by using successful human experience to guide online value learning and human actions to focus policy improvement. To this end, an independent implicit Q-learning (IQL) branch~\cite{iql} supplies a reference policy $\pi_{\mathrm{ref}}$ and action-value estimates $\mathbf I(s)$ for shaping the online critic $\mathbf Q_\theta$. The Human Action Fence then uses a learned score $H(s,a)$ to define the action neighborhood $\Omega_H(s)$ within which the actor $\pi_\phi$ searches for higher-value actions.

During interaction, a human may provide \emph{recovery}, a brief correction before returning control to $\pi_\phi$, or \emph{takeover}, continued human control to success. Both yield pre-intervention autonomous pairs $(s,a_{\mathrm{pre}})$ for $\mathcal D_{\mathrm{pre}}$, allowing local action values to be corrected relative to the reference. Takeovers additionally contribute successful human suffixes to $\mathcal D_{\mathrm{succ}}$ alongside successful demonstrations. To correct value overestimation at unrecoverable states, whole rollouts ending at human-marked unrecoverable states enter $\mathcal D_{\mathrm{fail}}$, and the marked states directly supervise both value learners. IQL value learning uses $\mathcal D_{\mathrm{IQL}}=\mathcal D_{\mathrm{succ}}\cup\mathcal D_{\mathrm{fail}}$, while only $\mathcal D_{\mathrm{succ}}$ is used to train $\pi_{\mathrm{ref}}$ and penalize online reference-action values that fall below the IQL estimates. Human-controlled transitions from demonstrations and both intervention types populate $\mathcal D_H$, whose state--action pairs train $H$, retaining neighborhood guidance from brief corrections as well as successful completions. All executed transitions enter replay $\mathcal D$. Following RLPD's sampling strategy~\cite{rlpd}, we update the online actor--critic with minibatches drawn equally from $\mathcal D$ and the human transition buffer $\mathcal D_H$, maintaining human guidance as autonomous experience accumulates.

\subsection{Human-Reference-Guided Value Shaping}
\label{sec:method-value-shaping}

To accelerate online value learning, we design an independent value reference based on IQL~\cite{iql}. IQL learns action-value vectors $\mathbf Q^I_j$ from $\mathcal D_{\mathrm{IQL}}$ using state-value backups, allowing successful human experience to provide value supervision for the online critic. Here $j\in\{1,2\}$ indexes the twin critics. $K$-dimensional value vectors are written in bold, with minima taken elementwise. Advantage-weighted imitation on $\mathcal D_{\mathrm{succ}}$ yields the reference policy $\pi_{\mathrm{ref}}$. We evaluate its action $a_{\mathrm{ref}}(s)=\pi_{\mathrm{ref}}(s)$ to obtain $\mathbf I(s)=\min_{j=1,2}\mathbf Q^I_j(s,a_{\mathrm{ref}}(s))$.

To retain this guidance while allowing improvement from online experience, we separate the value of the reference action from the relative advantage of other actions. For online critic $j$ with parameters $\theta_j$, $\mathbf B_j(s)$ represents the learnable values of the reference action. We parameterize relative action advantages by centering $\mathbf A_j(s,a)$ at $a_{\mathrm{ref}}(s)$, yielding the value before applying the action fence:
\begin{equation}
\begin{aligned}
    \Delta \mathbf A_j(s,a)
    &=\mathbf A_j(s,a)-\mathbf A_j(s,a_{\mathrm{ref}}(s)),\\
    \widetilde{\mathbf Q}_{\theta_j}(s,a)
    &=\mathbf B_j(s)+\Delta \mathbf A_j(s,a).
\end{aligned}
\label{eq:reference-centered-value}
\end{equation}
Thus, $\widetilde{\mathbf Q}_{\theta_j}(s,a_{\mathrm{ref}}(s))=\mathbf B_j(s)$. We use separate parameter sets for $\mathbf B$ and $\mathbf A$. On states from $\mathcal D_{\mathrm{succ}}$, we train this baseline against the IQL reference through a one-sided loss:
\begin{equation}
    \mathcal L_{\mathrm{floor}}
    =\mathbb E_{s\sim\mathcal D_{\mathrm{succ}}}\!\left[
        \frac{1}{2K}\sum_{j=1}^{2}
        \big\|[\operatorname{sg}(\mathbf I(s))-\mathbf B_j(s)]_+\big\|_2^2
    \right],
\label{eq:successful-reference-floor}
\end{equation}
where $[x]_+=\max(x,0)$ acts elementwise on vectors and $\operatorname{sg}$ denotes stop-gradient. Only estimates below the reference incur a penalty, leaving room for higher values supported by online experience.

Human interventions provide complementary feedback for correcting the values of preceding autonomous actions~\cite{pact}. We use this feedback to reduce their average advantage relative to the reference, with a margin $c>0$ equal to the step cost. For recorded pairs $(s,a_{\mathrm{pre}})\sim\mathcal D_{\mathrm{pre}}$, the basic ranking loss is
\begin{equation}
    \mathcal L_{\mathrm{corr}}
    =\mathbb E_{\mathcal D_{\mathrm{pre}}}\!\left[
        \frac12\sum_{j=1}^{2}
        \left[c+\frac1K\mathbf 1^\top
        \Delta\mathbf A_j(s,a_{\mathrm{pre}})\right]_+
    \right].
\label{eq:intervention-correction}
\end{equation}
Here $\mathbf 1\in\mathbb R^K$ is the all-ones vector. With $a_{\mathrm{ref}}$ fixed in this loss, the correction adjusts $\Delta\mathbf A$ while preserving the reference baseline $\mathbf B$.

Explicit annotations of unrecoverable states provide direct supervision for this baseline. Assuming no subsequent success and a cost $c$ per step, we model the $h$-step return as
\begin{equation}
    F_h=-c\sum_{k=0}^{h-1}\gamma^k,\qquad F_0=0.
\label{eq:unrecoverable-value}
\end{equation}
We fit the IQL values and $\widetilde{\mathbf Q}_{\theta_j}$ at these states to $\mathbf F=(F_1,\ldots,F_K)^\top$ using squared-error losses. This directly calibrates the reference baseline, mitigating overestimation that relative action correction alone cannot resolve.

\subsection{Human Action Fence}
\label{sec:method-action-fence}

\begin{figure*}[t]
    \centering
    \includegraphics[width=\textwidth]{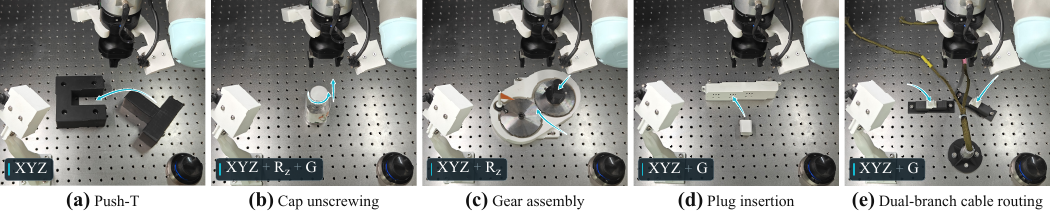}
    \caption{Real-world manipulation tasks. In-image labels indicate controlled action dimensions: $XYZ$ denotes Cartesian translation, $R_z$ rotation about the end-effector $z$-axis, and $G$ gripper actuation. Arrows illustrate task motions or indicate manipulation locations.}
    \label{fig:manipulation-tasks}
\end{figure*}

\begin{figure*}[t]
    \centering
    \includegraphics[width=\textwidth]{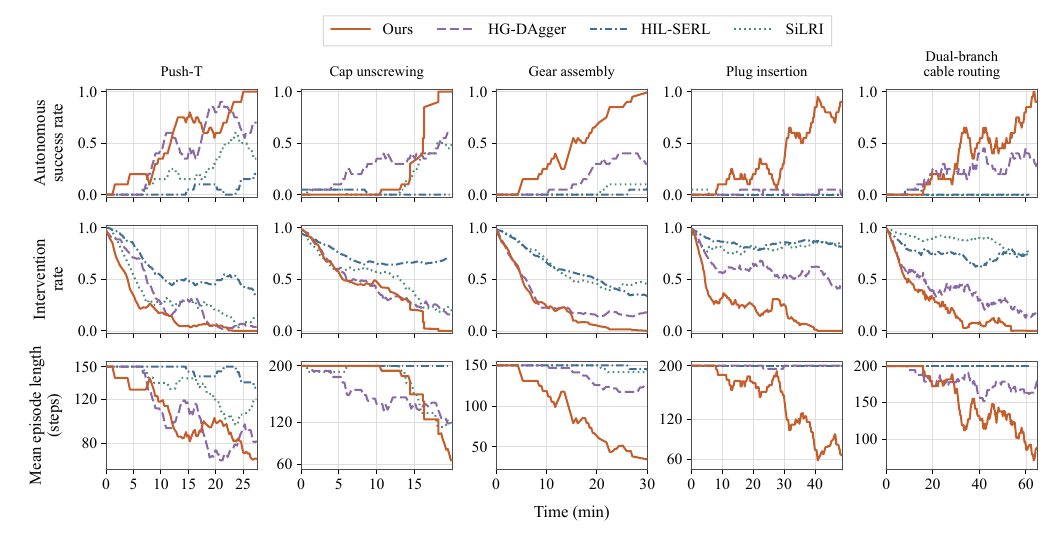}
    \caption{Online learning performance of ReF-HIL (Ours), HG-DAgger, HIL-SERL, and SiLRI. Columns correspond to the five manipulation tasks; rows show autonomous success rate, intervention rate, and mean episode length. Curves show 20-episode moving averages against accumulated active training time.}
    \label{fig:comparison-curves}
\end{figure*}

We design the Human Action Fence to improve learning efficiency by concentrating policy optimization within a learned neighborhood of human actions. Using $(s,a_H)$ from $\mathcal D_H$, we regress a nonnegative score $H(s,a)$ to Gaussian kernel targets for sampled query actions $a$,
\begin{equation}
    \kappa_\sigma(a,a_H)
    =(2\pi\sigma^2)^{-d_a/2}
    \exp\!\left(-\frac{\|a-a_H\|_2^2}{2\sigma^2}\right),
\label{eq:human-action-kernel}
\end{equation}
where $d_a$ is the action dimension and $\sigma$ sets the bandwidth in normalized action coordinates; each kernel has unit integral over $\mathbb R^{d_a}$. For $0<\rho\le1$, we define
\begin{equation}
    \Omega_H(s)=\left\{a\in\mathcal A:
    H(s,a)\ge\rho H(s,a_{\mathrm{ref}}(s))\right\}.
\label{eq:human-action-neighborhood}
\end{equation}
The relative threshold ensures that the reference action remains inside the neighborhood.

Within this neighborhood, we retain the learned relative advantage $\Delta \mathbf A_j(s,a)$. Outside it, we replace this advantage with a distance penalty relative to the same reference-action baseline:
\begin{equation}
    \mathbf Q_{\theta_j}(s,a)=
    \begin{cases}
        \widetilde{\mathbf Q}_{\theta_j}(s,a)&\!\!a\in\Omega_H(s),\\
        \mathbf B_j(s)-\bigl(c+\|a-a_{\mathrm{ref}}(s)\|_2^2\bigr)\mathbf 1
        &\!\!a\notin\Omega_H(s).
    \end{cases}
\label{eq:fenced-action-value}
\end{equation}
Values assigned outside $\Omega_H(s)$ are componentwise below $\mathbf B_j(s)$, ensuring that these actions cannot outrank the reference through value extrapolation.

We optimize the policy with the value objective
\begin{equation}
    \mathcal L_\pi^Q
    =-\mathbb E_s\!\left[
        \frac1K\mathbf 1^\top
        \min_{j=1,2}\mathbf Q_{\theta_j}(s,\pi_\phi(s))
    \right].
\label{eq:fenced-policy-objective}
\end{equation}
Within $\Omega_H(s)$, learned value gradients can favor actions that differ from the reference. Outside $\Omega_H(s)$, the quadratic term provides an action gradient toward the reference. This supports local improvement without an action-matching penalty inside the neighborhood.

To train the online critic, we construct TD targets by comparing policy and reference actions at the next state, following IBRL~\cite{ibrl}. Using target critics with parameters $\bar\theta_j$, we define the continuation values $C_h$ and target components $y_h$ as
\begin{equation}
\begin{aligned}
    C_h(s')&=\max_{a'\in\{\pi_\phi(s'),a_{\mathrm{ref}}(s')\}}
        \min_{j=1,2}Q_{\bar\theta_j,h}(s',a'),\\
    y_h&=r+\gamma(1-d)C_{h-1}(s'),\qquad C_0=0.
\end{aligned}
\label{eq:reference-policy-backup}
\end{equation}
For explicitly unrecoverable next states, $F_{h-1}$ replaces $C_{h-1}(s')$. We fit $\widetilde{\mathbf Q}_{\theta_j}$ to the target vector $\mathbf y$ on transitions accepted by the fence, retaining transitions into annotated unrecoverable states even when their actions are rejected. Besides, actions rejected by the fence receive an auxiliary ranking penalty on their relative advantages. These updates propagate higher estimated returns and annotated failure costs to preceding actions.

\section{Experiments}
\label{sec:experiments}

\subsection{Experimental Setup}
\label{sec:experimental-setup}

\subsubsection{Implementation Details}
\label{sec:implementation-details}
We conduct experiments on a UR5e robotic arm equipped with a Robotiq Hand-E gripper. RGB images from two wrist-mounted Intel RealSense D405 cameras and one global D405 camera are encoded by a frozen ImageNet-pretrained ResNet-18. The visual features are combined with end-effector velocities, forces and torques, gripper opening, and gravity direction in the end-effector frame to form the policy input. The policy generates Cartesian motion commands at 10~Hz, which are executed through an admittance controller. To support online learning, we use an asynchronous actor--learner architecture, with the actor running on an AMD Radeon RX 9070 XT (16G) and the learner on an NVIDIA GeForce RTX 5090 (32G).

For each task, training begins with 20 successful human demonstration episodes and continues through autonomous interaction with human corrections provided through a SpaceMouse. Each episode starts from an end-effector position randomly sampled within task-specific bounds. Pre-intervention correction uses up to 10 consecutive autonomous state-action pairs immediately preceding each human intervention. The relative threshold of the Human Action Fence is set to $\rho=0.9$. Experimental settings and hyperparameters are detailed in the supplementary video.

\subsubsection{Task Definitions}
\label{sec:experiment-tasks}
\begin{table*}[t]
    \caption{Performance across the five manipulation tasks}
    \label{tab:performance-comparison}
    \centering
    \begin{minipage}{\textwidth}
        \centering
        \fontsize{8}{10}\selectfont
\setlength{\tabcolsep}{0pt}
\renewcommand{\arraystretch}{1.15}
\begin{tabular}{@{}p{.145\textwidth}*{10}{>{\centering\arraybackslash}p{.0855\textwidth}}@{}}
\toprule
\multirow[c]{2}{*}[-2.4pt]{Method} & \multicolumn{2}{c}{\parbox[c]{.171\textwidth}{\centering Push-T}} & \multicolumn{2}{c}{\parbox[c]{.171\textwidth}{\centering Cap unscrewing}} & \multicolumn{2}{c}{\parbox[c]{.171\textwidth}{\centering Gear assembly}} & \multicolumn{2}{c}{\parbox[c]{.171\textwidth}{\centering Plug insertion}} & \multicolumn{2}{c}{\parbox[c]{.171\textwidth}{\centering Dual-branch\\cable routing}} \\
\cmidrule(lr){2-3}\cmidrule(lr){4-5}\cmidrule(lr){6-7}\cmidrule(lr){8-9}\cmidrule(lr){10-11}
& SR (\%) $\uparrow$ & Length $\downarrow$ & SR (\%) $\uparrow$ & Length $\downarrow$ & SR (\%) $\uparrow$ & Length $\downarrow$ & SR (\%) $\uparrow$ & Length $\downarrow$ & SR (\%) $\uparrow$ & Length $\downarrow$ \\
\midrule
HG-DAgger~\cite{hgdagger} & $86.7\!\pm\!2.9$ & $\boldsymbol{66.6\!\pm\!1.8}$ & $76.7\!\pm\!2.9$ & $109.8\!\pm\!7.4$ & $38.3\!\pm\!2.9$ & $118.4\!\pm\!1.5$ & $5.0\!\pm\!0.0$ & $195.2\!\pm\!0.0$ & $5.0\!\pm\!0.0$ & $193.4\!\pm\!0.0$ \\
HIL-SERL~\cite{hilserl} & $16.7\!\pm\!2.9$ & $134.1\!\pm\!2.8$ & $1.7\!\pm\!2.9$ & $198.5\!\pm\!2.7$ & $6.7\!\pm\!2.9$ & $143.9\!\pm\!2.3$ & $0.0\!\pm\!0.0$ & $200.0\!\pm\!0.0$ & $0.0\!\pm\!0.0$ & $200.0\!\pm\!0.0$ \\
SiLRI~\cite{silri} & $55.0\!\pm\!5.0$ & $100.3\!\pm\!2.8$ & $36.7\!\pm\!5.8$ & $138.5\!\pm\!9.9$ & $10.0\!\pm\!0.0$ & $141.5\!\pm\!0.0$ & $0.0\!\pm\!0.0$ & $200.0\!\pm\!0.0$ & $0.0\!\pm\!0.0$ & $200.0\!\pm\!0.0$ \\
\midrule
\textbf{ReF-HIL (Ours)} & $\boldsymbol{100.0\!\pm\!0.0}$ & $69.3\!\pm\!2.9$ & \hspace*{-1pt}$\boldsymbol{100.0\!\pm\!0.0}$\hspace*{1pt} & \hspace*{1pt}$\boldsymbol{42.4\!\pm\!1.4}$\hspace*{-1pt} & $\boldsymbol{95.0\!\pm\!5.0}$ & $\boldsymbol{40.1\!\pm\!5.6}$ & $\boldsymbol{91.7\!\pm\!2.9}$ & $\boldsymbol{64.4\!\pm\!5.0}$ & $\boldsymbol{95.0\!\pm\!5.0}$ & $\boldsymbol{78.3\!\pm\!7.0}$ \\
\bottomrule
\end{tabular}\par
\vspace{4pt}
\raggedright\fontsize{8}{9.5}\selectfont
All results are obtained from separate evaluations of final policies without human intervention. SR: autonomous success rate. Length (steps): batch mean of actual step counts for autonomous successes and task horizons for other episodes (150 for Push-T and Gear assembly; 200 for the other tasks). Values are mean $\pm$ sample SD across batches, not independent training runs.\par

    \end{minipage}
\end{table*}

\begin{figure}[t]
    \centering
    \includegraphics[width=\columnwidth]{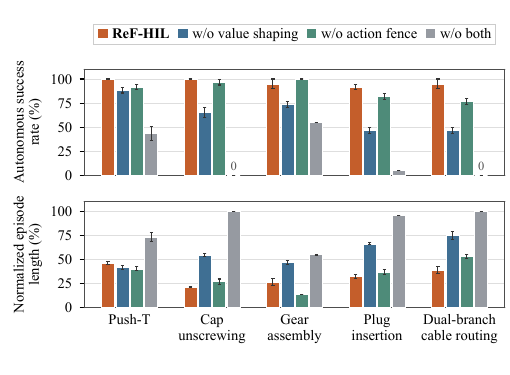}
    \caption{Ablation results across five manipulation tasks, obtained from separate evaluations of final policies without human intervention. Bars show autonomous success rate (top) and episode length normalized by the task horizon (bottom), using the Length definition in Table~\ref{tab:performance-comparison}. Error bars denote sample standard deviations across batches.}
    \label{fig:ablation-results}
\end{figure}

We evaluate ReF-HIL on five diverse and challenging real-world manipulation tasks (Fig.~\ref{fig:manipulation-tasks}), spanning contact-constrained motion, precision assembly, and long-horizon manipulation of deformable objects.
\begin{itemize}[\setlength{\labelindent}{0pt}\setlength{\labelsep}{0.5em}\setlength{\topsep}{2pt}\setlength{\itemsep}{1pt}]
    \item \textbf{Push-T:} Pushing a T-shaped block into its target pose.
    \item \textbf{Cap unscrewing:} Grasping and unscrewing a bottle cap through thread-constrained motion until it is detached.
    \item \textbf{Gear assembly:} Installing a grasped gear by inserting its shaft into a hole and meshing it with the adjacent gear.
    \item \textbf{Plug insertion:} Picking up a plug from the table and inserting it into a power-strip socket.
    \item \textbf{Dual-branch cable routing:} Grasping and routing two deformable cable branches into their designated slots in a fixed order.
\end{itemize}

\subsubsection{Baselines}
\label{sec:experiment-baselines}
To assess how effectively ReF-HIL uses human guidance, we compare it with the following methods on the five tasks described above:
\begin{itemize}[\setlength{\labelindent}{0pt}\setlength{\labelsep}{0.5em}\setlength{\topsep}{2pt}\setlength{\itemsep}{1pt}]
    \item \textbf{HG-DAgger}~\cite{hgdagger} learns a policy by imitating demonstrations and aggregated human corrections.
    \item \textbf{HIL-SERL}~\cite{hilserl} combines demonstrations and online interventions in replay buffers for efficient off-policy RL.
    \item \textbf{SiLRI}~\cite{silri} adjusts imitation constraints according to intervention uncertainty, with state-wise Lagrange multipliers balancing human guidance and policy improvement.
\end{itemize}
We use the publicly available implementations of all three baselines, adapted to our robotic platform. For each task, all methods use the same 20 demonstrations, observation and action spaces, initial-state randomization, and evaluation protocol. They also use the same pretraining budget and online update schedule, with effective update rates limited by computational throughput.

\subsubsection{Evaluation Metrics}
\label{sec:experiment-metrics}
We evaluate online learning in terms of autonomous success, human intervention, and execution efficiency. \textbf{Autonomous success rate} measures the fraction of episodes in which the task is successfully completed without human intervention. \textbf{Intervention rate} is the mean per-episode fraction of human-controlled steps. \textbf{Mean episode length} is computed using the actual step count for successful episodes without human intervention and the maximum allowed step count for all other episodes. In Fig.~\ref{fig:comparison-curves}, these metrics are plotted as 20-episode moving averages against accumulated active training time, which excludes pretraining.
Table~\ref{tab:performance-comparison} and Fig.~\ref{fig:ablation-results} report results from three batches of 20 episodes each.

\subsection{Comparative Experiments}
\label{sec:comparative-experiments}
\begingroup
\clubpenalty=10000
\widowpenalty=10000
We compare ReF-HIL with three baselines on five real-world manipulation tasks, with results summarized in Fig.~\ref{fig:comparison-curves} and Table~\ref{tab:performance-comparison}. Overall, ReF-HIL achieves higher autonomous success rates with less reliance on human intervention than the baselines.

On Push-T, it reaches 100\% autonomous success after approximately 25 minutes of active training. Although HG-DAgger also learns quickly, ReF-HIL achieves a higher success rate (100\% versus 86.7\%) with a slightly higher penalized episode length (69.3 versus 66.6 steps). This slight increase reflects additional corrective pushes that adjust the block's pose and help improve policy robustness.
On cap unscrewing, ReF-HIL reaches 100\% autonomous success after approximately 18 minutes of active training. Compared with HG-DAgger, the best-performing baseline on this task, it achieves both a higher success rate (100\% versus 76.7\%) and a lower penalized episode length (42.4 versus 109.8 steps). Here, successful completion requires following a thread-constrained path until the cap detaches. By learning from successful human trajectories, ReF-HIL's independent IQL reference provides value guidance for intermediate actions along this path before the online policy can reliably complete the task.
On gear assembly, ReF-HIL achieves a 95.0\% success rate, compared with 38.3\% for HG-DAgger, the strongest baseline on this task, while reducing the penalized episode length from 118.4 to 40.1 steps.

The performance gap widens on plug insertion, where ReF-HIL achieves 91.7\% success while all baselines remain at 5\% or below. It reaches 90\% autonomous success after approximately 41 minutes of active training, with intervention subsequently falling to zero. Plug insertion requires sequential grasping and contact-rich insertion. Sparse success feedback complicates credit assignment across these stages, while imitation-based methods may reproduce suboptimal human corrections that hinder reliable completion.

Finally, dual-branch cable routing tests long-horizon manipulation, requiring a single policy to complete two successive grasp-and-route stages. Errors can accumulate across stages, while success feedback is available only after both branches are seated. Despite these challenges, ReF-HIL reaches 90\% autonomous success after approximately 62 minutes of active training and achieves a success rate of 95.0\%, compared with at most 5\% for the baselines.

\par
\endgroup

\subsection{Ablation Studies}
\label{sec:ablation-studies}
We compare ReF-HIL with three ablations in Fig.~\ref{fig:ablation-results}: \textbf{w/o value shaping}, \textbf{w/o action fence}, and \textbf{w/o both}, which remove human-reference value shaping, the Human Action Fence, and both modules, respectively. All variants retain the IQL reference policy and use its actions as candidates in TD~backups.

Retaining either component improves success rates over \textbf{w/o both} on all five tasks. Removing value shaping reduces success on every task, with the largest decrease on long-horizon cable routing, from 95.0\% to 46.7\%, accompanied by an increase in normalized episode length from 39.1\% to 74.7\%. Action constraints alone therefore do not recover the completion reliability achieved with value shaping.

The action fence further improves success on Push-T, cap unscrewing, plug insertion, and dual-branch cable routing, with the largest gain on cable routing (18.3 percentage points). On Push-T, the full method achieves higher success than either single-component variant despite a slightly longer normalized episode length. On gear assembly, however, removing the fence increases success from 95.0\% to 100\% and reduces normalized episode length from 26.7\% to 13.5\%. For tasks requiring only local adjustments, a human-action constraint may offer limited exploration benefits while still restricting policy improvement, particularly when human corrections are inconsistent. The fence's contribution is therefore task dependent.

\Needspace{5\baselineskip}
\subsection{Mechanism Analysis}
\label{sec:mechanism-analysis}
\subsubsection{Value Learning}
\label{sec:value-learning-analysis}
\begin{figure}[t]
    \centering
    \includegraphics[width=\columnwidth]{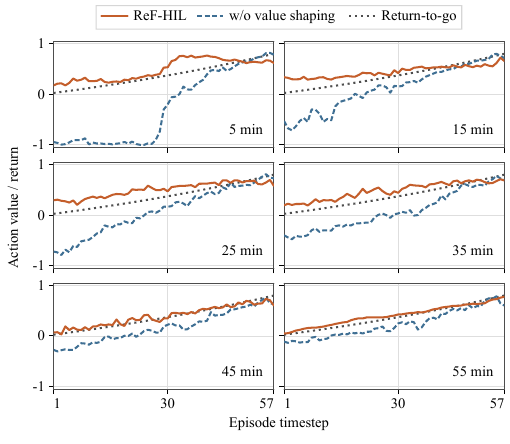}
    \caption{Raw twin-min critic estimates and recorded discounted return-to-go along the same selected 57-step autonomous cable-routing trajectory. Both critics use the remaining episode-budget horizon. Panels show approximate time after pretraining; the orange, blue, and grey curves denote ReF-HIL, w/o value shaping, and recorded return, respectively.}
    \label{fig:value-inspection}
\end{figure}

To examine how value shaping supports long-horizon learning, we evaluate saved critic checkpoints on identical state--action pairs from an autonomous cable-routing trajectory selected from late training. This trajectory completes the task in 57 steps, compared with 61 steps for the shortest of the 20 initial demonstrations. Fig.~\ref{fig:value-inspection} compares the action-value estimates of ReF-HIL and w/o value shaping at different training stages with the trajectory's recorded discounted return-to-go.

\begingroup
\widowpenalty=10000
At approximately 5 minutes, the critic without value shaping assigns reasonable values near task completion but substantially underestimates earlier state--action pairs. ReF-HIL alleviates this underestimation, achieving a mean absolute deviation from the recorded return of 0.136, compared with 0.648 for the ablation. Although ReF-HIL initially overestimates parts of the trajectory, these deviations diminish during subsequent online learning, and its estimates closely track the recorded return at 45--55 minutes. In contrast, underestimation persists over much of the earlier trajectory without value shaping. This pattern is consistent with human-derived value references providing early guidance for states far from task completion, followed by refinement through online experience.
\par
\endgroup

\subsubsection{Policy Improvement}
\label{sec:policy-improvement-action-fence}
\begin{figure}[t]
    \centering
    \includegraphics[width=\columnwidth]{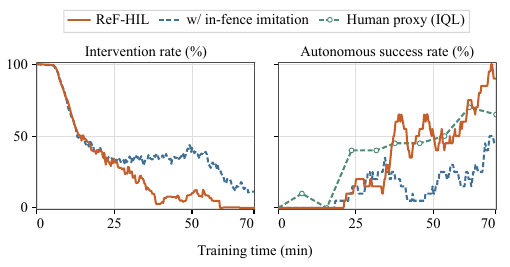}
    \caption{Effect of in-fence imitation on dual-branch cable routing. Both online variants retain the Human Action Fence; the blue dashed variant adds an $L_2$ imitation penalty toward the reference action inside the neighborhood. Orange and blue curves show 20-episode moving averages from one training run per variant, with a common cutoff of approximately 70 minutes. Green markers show 20-trial checkpoint evaluations of the IQL reference policy from a separate run. Training time excludes pretraining.}
    \label{fig:in-fence-imitation}
\end{figure}

\begingroup
\widowpenalty=10000
We investigate how continuous imitation within the Human Action Fence affects policy improvement on dual-branch cable routing. The comparison variant retains the fence and adds an $L_2$ penalty toward the reference action inside the learned neighborhood, with a 1:1 weighting between the RL and imitation losses. We also evaluate the IQL reference policy learned from human data, using 20 trials at each checkpoint.
\par
\endgroup

Fig.~\ref{fig:in-fence-imitation} shows similar early reductions in intervention for the two online variants, followed by a divergence in autonomous performance. By approximately 70 minutes, ReF-HIL achieves 90\% autonomous success with zero intervention, whereas the in-fence imitation variant reaches 45\% success and retains an intervention rate of 11.3\%. The independently evaluated IQL reference policy achieves 65\% success at its final checkpoint.

Inside the fence, the $L_2$ imitation penalty still assigns a cost to deviating from the reference action, even when such a deviation increases the estimated return. The actor therefore trades off value improvement against reference proximity despite remaining within the permitted neighborhood. The observed divergence is consistent with this constraint limiting later policy refinement. ReF-HIL separates these roles: the human-action neighborhood constrains exploration, while value estimates guide optimization within it, leaving room to improve upon the learned reference.

\section{Conclusion}
\label{sec:conclusion}

We presented ReF-HIL, an efficient human-in-the-loop reinforcement learning framework. Human-Reference-Guided Value Shaping provides an independent value reference for online learning, while the Human Action Fence focuses policy optimization within a learned human-action neighborhood without imposing imitation penalties inside it. Across five diverse and challenging real-world manipulation tasks, ReF-HIL demonstrates improved overall learning efficiency and higher autonomous success rates than the evaluated baselines. It reaches 90\% autonomous success in 18--63 minutes of active training and achieves final success rates of 91.7--100\%. These findings support using human experience to accelerate value learning while preserving value-driven policy improvement within a learned action neighborhood.

The effectiveness of the Human Action Fence varies across tasks, suggesting the need for more adaptive constraints. Future work will investigate fences that account for the reliability of human guidance and online learning progress. We will also explore adapting ReF-HIL for efficient post-training of vision-language-action (VLA) models using real-world interaction and human feedback.

\bibliographystyle{IEEEtran}
\begingroup
\renewcommand{\baselinestretch}{0.94}
\bibliography{IEEEfull,reference_style,references}
\endgroup

\end{document}